\documentclass[twoside,11pt]{article}
\usepackage{jagi}
\usepackage[utf8]{inputenc}
\usepackage{natbib}
\usepackage{amssymb}
\usepackage{amsmath}
\usepackage{mathrsfs}
\usepackage{mathdots}

\jagiheading{12}{1}{1--25}{2021}{2020-10-16}{2021-01-08}{10.2478/jagi-2021-0001}{S.\ Alexander \& B.\ Hibbard}

\ShortHeadings{Measuring Intelligence and Growth Rate}{S.\ Alexander \& B.\ Hibbard}

\title{Measuring Intelligence and Growth Rate: Variations on
Hibbard's Intelligence Measure}

\author{\name Samuel Alexander \email samuelallenalexander@gmail.com \\
    \addr (Primary Author)\\
    The U.S.\ Securities and Exchange Commission, New York Regional Office\\
    \AND
    \name Bill Hibbard \\
    \addr Space Science and Engineering Center, University of Wisconsin, Madison, WI
}

\editor{Marcus Hutter}

\begin{document}

\maketitle

\begin{abstract}
    In 2011, Hibbard suggested an intelligence measure for agents
    who compete in an adversarial sequence prediction game. We argue
    that Hibbard's idea should actually be considered as two separate
    ideas: first, that the intelligence of such agents can be measured
    based on the growth rates of the runtimes of the competitors that
    they defeat; and second, one specific (somewhat arbitrary) method for measuring said
    growth rates. Whereas Hibbard's intelligence measure is based on the latter
    growth-rate-measuring method, we survey
    other methods for measuring function
    growth rates, and exhibit the resulting Hibbard-like intelligence measures
    and taxonomies. Of particular interest, we obtain intelligence taxonomies
    based on Big-O and Big-Theta notation systems, which taxonomies
    are novel in that they challenge conventional notions of what an
    intelligence measure should look like. We discuss how intelligence measurement
    of sequence predictors can indirectly serve as intelligence measurement for
    agents with Artificial General Intelligence (AGIs).
\end{abstract}

\section{Introduction}

In his insightful paper, \citet{hibbard} introduces a novel
intelligence measure (which we will here refer to as the \emph{original Hibbard measure})
for agents who play a game of adversarial sequence prediction
\citep{hibbard2008adversarial}
``against a hierarchy of increasingly difficult sets of'' evaders (environments that attempt
to emit $1$s and $0$s in such a way as to evade prediction).
The levels of Hibbard's hierarchy are labelled by natural numbers, and
an agent's original Hibbard measure is the maximum $n\in\mathbb N$ such that
said agent learns to predict all the evaders in the $n$th level of the hierarchy,
or implicitly\footnote{Hibbard does not explicitly include the $\infty$ case in his
definition, but in his Proposition 3 he refers to agents having ``finite intelligence'', and
it is clear from context that by this he means agents who fail to predict some evader
somewhere in the hierarchy.} an agent's original Hibbard measure is $\infty$
if said agent learns to predict all the evaders in all levels of Hibbard's hierarchy.

The hierarchy which Hibbard uses to measure intelligence is based on the growth
rates of the runtimes of evaders.
We will argue that Hibbard's idea is really a combination of two
orthogonal ideas. First: that in some sense the intelligence of a predicting agent
can be measured based on the growth rates of the runtimes of the evaders whom that
predictor learns to predict. Second: Hibbard proposed one particular method for
measuring said growth rates. The growth rate measurement which Hibbard proposed yields
a corresponding intelligence measure for these agents. We will argue that \emph{any}
method for measuring growth rates of functions yields a corresponding
\emph{adversarial sequence prediction intelligence} measure (or \emph{ASPI} measure
for short) provided the underlying number system provides a way of choosing canonical bounds
for bounded sets. If the underlying number system does not provide a way of choosing
canonical bounds for bounded sets, the growth-rate-measure will yield a corresponding
ASPI taxonomy (like the big-$O$ taxonomy of asymptotic complexity).

The particular method which Hibbard used to measure function growth rates is
not very standard. We will survey other
ways of measuring function growth rates,
and these will yield corresponding ASPI measures and taxonomies.

The structure of the paper is as follows.
\begin{itemize}
    \item
    In Section \ref{originalmeasuresection}, we review the original Hibbard measure.
    \item
    In Section \ref{growthratesection}, we argue that any method of measuring
    growth rates of functions yields an ASPI measure or taxonomy,
    and that the original Hibbard measure is just a special case resulting from
    one particular method of measuring function growth rate.
    \item
    In Section \ref{bigosection}, we consider Big-O notation and Big-$\Theta$ notation
    and define corresponding ASPI taxonomies.
    \item
    In Section \ref{majorizationsection}, we consider solutions to the problem of
    measuring growth rates of functions using majorization hierarchies, and define
    corresponding ASPI measures.
    \item
    In Section \ref{hyperrealsurrealsection}, we consider solutions to the
    problem of measuring growth rates of functions using more abstract number
    systems, namely the hyperreal numbers and the surreal numbers. We do not
    assume previous familiarity with these number systems.
    \item
    In Section \ref{prosandconssection}, we give pros and cons of different
    ASPI measures and taxonomies.
    \item
    In Section \ref{conclusionsection}, we summarize and make concluding remarks.
\end{itemize}

\section{Hibbard's original measure}
\label{originalmeasuresection}

Hibbard proposed an intelligence measure for measuring the intelligence
of agents who compete to predict evaders in a game of
adversarial sequence prediction (we define this
formally below). A predictor $p$ (whose intelligence we want to measure)
competes against evaders $e$. In each step of the game,
both predictor and evader simultaneously choose a binary digit, $1$ or $0$.
Only after both of them have made their choice do they see which choice the other
one made, and then the game proceeds to the next step. The predictor's goal in
each round is to choose the same digit that the evader will choose;
the evader's goal is to choose a different digit than the predictor. The predictor
wins the game (and is said to \emph{learn to predict $e$}, or simply to
\emph{learn $e$}) if, after finitely many
initial steps, eventually the predictor always chooses the same digit as the
evader.

\begin{definition}
By $B$, we mean the binary alphabet $\{0,1\}$. By $B^*$, we mean the set of all
finite binary sequences. By $\langle\rangle$ we mean the empty binary sequence.
\end{definition}

\begin{definition}
\label{evaderpredictordefn}
    (Predictors and evaders)
    \begin{enumerate}
        \item
        By a \emph{predictor}, we mean a Turing machine\footnote{The measures we introduce
        in this paper would also work if we defined predictors as not-necessarily-computable
        functions $B^*\to B$, but this would not add much insight. We prefer
        to emphasize the duality between predictors and evaders when each is a Turing
        machine.} $p$
        which takes as input a finite (possibly empty) binary sequence
        $(x_1,\ldots,x_n)\in B^*$
        (thought of as a sequence of \emph{evasions})
        and outputs $0$ or $1$ (thought of as a \emph{prediction}), which output
        we write as $p(x_1,\ldots,x_n)$.
        \item
        By an \emph{evader}, we mean a Turing machine $e$
        which takes as input a finite (possibly empty) binary sequence
        $(y_1,\ldots,y_n)\in B^*$
        (thought of as a sequence of \emph{predictions})
        and outputs $0$ or $1$ (thought of as an \emph{evasion}), which output
        we write as $e(y_1,\ldots,y_n)$.
        \item
        For any predictor $p$ and evader $e$, the \emph{result of $p$ playing the
        game of adversarial sequence
        prediction against $e$} (or more simply, the \emph{result of
        $p$ playing against $e$}) is the infinite binary sequence
        $(x_1,y_1,x_2,y_2,\ldots)$
        defined as follows:
        \begin{enumerate}
            \item
            The first evasion
            $x_1=e(\langle\rangle)$ is
            the output of $e$ when run on the empty prediction-sequence.
            \item
            The first prediction
            $y_1=p(\langle\rangle)$ is
            the output of $p$ when run on the empty evasion-sequence.
            \item
            For all $n>0$, the $(n+1)$th evasion
            $x_{n+1}=e(y_1,\ldots,y_n)$ is
            the output of $e$ on the sequence of the first $n$ predictions.
            \item
            For all $n>0$, the $(n+1)$th prediction
            $y_{n+1}=p(x_1,\ldots,x_n)$ is
            the output of $p$ on the sequence of the first $n$ evasions.
        \end{enumerate}
        \item
        Suppose $r=(x_1,y_1,x_2,y_2,\ldots)$ is the result of a predictor $p$ playing
        against an evader $e$. For every $n\geq 1$,
        we say \emph{the predictor wins round $n$ in $r$}
        if $x_n=y_n$; otherwise,
        \emph{the evader wins round $n$ in $r$}.
        We say that \emph{$p$ learns to predict $e$}
        (or simply that \emph{$p$ learns $e$}) if there is some $N\in\mathbb N$
        such that for all $n>N$, $p$ is the winner of round $n$ in $r$.
    \end{enumerate}
\end{definition}

Note that if $e$ simply ignores its inputs $(y_1,\ldots,y_n)$ and instead
computes $e(y_1,\ldots,y_n)$ based only on $n$, then $e$ is essentially a sequence.
Thus Definition \ref{evaderpredictordefn} is a generalization of sequence prediction,
which many authors have written about (such as \citet{legg2006there}, who gives many
references). In the future, it could be interesting to consider variations of
the game involving probability in various ways, for example, where the predictor wins
if his guesses have $>50\%$ win rate, or where the predictor states how confident he is
about each guess, or other such variations.

In the following definition, we differ from Hibbard's original paper
because of a minor (and fortunately, easy-to-fix) error there.

\begin{definition}
\label{tsubedefinition}
    Suppose $e$ is an evader.
    For each $n\in\mathbb N$, let $t_e(n)$ be the maximum number of steps that $e$ takes
    to run on any length-$n$ sequence of binary digits.
    In other words, $t_e(0)$ is the number of steps $e$ takes to run on $\langle\rangle$,
    and for all $n>0$,
    \[
        t_e(n) = \max_{b_1,\ldots,b_n\in \{0,1\}}
        (\text{number of steps $e$ takes to run on $(b_1,\ldots,b_n)$}).
    \]
\end{definition}

\begin{example}
    Let $e$ be an evader. Then
    $t_e(2)$ is equal to the number of steps $e$ takes to run on input
    $(0,0)$, or to run on input $(0,1)$, or to run on input $(1,0)$, or to run on input
    $(1,1)$---whichever of these four possibilities is largest.
\end{example}

\begin{definition}
\label{functionsuccdefn}
    Suppose $f:\mathbb N\to\mathbb N$ and $g:\mathbb N\to\mathbb N$.
    We say $f$ \emph{majorizes} $g$, written
    $f\succ g$, if there is some $n_0\in \mathbb N$ such that for all
    $n>n_0$, $f(n)>g(n)$.
\end{definition}

\begin{definition}
\label{evadersetdefinition}
    Suppose $f:\mathbb N\to\mathbb N$. We define
    $E_f$ to be the set of all evaders $e$ such that
    $f\succ t_e$.
\end{definition}

\begin{definition}
\label{classichibbardmeasuredefn}
    (The original Hibbard measure)
    Let $g_1,g_2,\ldots$ be the enumeration of the primitive recursive
    functions given by \citet{liu1960enumeration}.
    For each $m>0$, define $f_m:\mathbb N\to\mathbb N$ by
    \[f_m(k)=\max_{0<i\leq m}\max_{j\leq k}g_i(j).\]
    For any predictor $p$, we define the \emph{original Hibbard intelligence of $p$}
    to be the maximum $m>0$
    such that $p$ learns to predict $e$ for every $e\in E_{f_m}$
    (or $0$ if there is no such $m$, or $\infty$ if $p$ learns to predict $e$
    for every $e\in E_{f_m}$ for every $m>0$).
\end{definition}

The following result shows that the measure in
Definition \ref{classichibbardmeasuredefn} does not overshoot the agents being
measured.

\begin{proposition}
\label{complicatedproposition}
    For any integer $m\geq 1$, there is a predictor $p$ with original Hibbard
    measure $\geq m$.
\end{proposition}

\begin{proof}
    This is part of Proposition 3 of \citet{hibbard}, but we give a self-contained
    proof here because we will state similar results about other measures below.

    For each computable $f:\mathbb N\to\mathbb N$,
    let $p_f$ be the predictor who proceeds as follows when given
    evasion-sequence $(x_1,\ldots,x_n)\in B^*$ as input.

    First, by calling itself recursively on inputs
    \[\langle\rangle, (x_1), (x_1,x_2), \ldots, (x_1,\ldots,x_{n-1}),\]
    $p_f$ determines the prediction-sequence $(y_1,\ldots,y_n)$ as in
    Definition \ref{evaderpredictordefn}.

    Next, $p_f$ considers the first $n$ Turing machines, $T_1,\ldots,T_n$.
    For $1\leq i\leq n$, say that $T_i$ is an \emph{$n$th-order $e$-lookalike} if
    the following requirements hold:
    \begin{itemize}
        \item If $T_i$ halts in $\leq f(0)$ steps on input
        $\langle\rangle$, then $T_i$ outputs $x_1$ on that input.
        \item If $T_i$ halts in $\leq f(1)$ steps on input
        $(y_1)$, then $T_i$ outputs $x_2$ on that input.
        \item If $T_i$ halts in $\leq f(2)$ steps on input
        $(y_1,y_2)$, then $T_i$ outputs $x_3$ on that input.
        \item $\ldots$
        \item If $T_i$ halts in $\leq f(n-1)$ steps on input
        $(y_1,\ldots,y_{n-1})$, then $T_i$ outputs $x_n$ on that input.
        \item $T_i$ halts in $\leq f(n)$ steps on input
        $(y_1,\ldots,y_n)$ with some output $X_{i,n}$.
    \end{itemize}
    By simulating $T_1,\ldots,T_n$ as needed (which only requires finitely many steps),
    $p_f$ determines if any $T_i$ is an $n$th-order $e$-lookalike.
    If so, $p_f$ outputs $p_f(x_1,\ldots,x_n)=X_{i,n}$ for the minimal such $i$.
    If not, $p_f$ outputs $0$.

    \textbf{Claim:} For every computable $f:\mathbb N\to\mathbb N$,
    for every evader $e$,
    if $(x_1,y_1,x_2,y_2,\ldots)$ is the result of $p_f$ playing against $e$,
    then for all-but-finitely-many $n\in\mathbb N$,
    if $f(n)>t_e(n)$ then $x_{n+1}=y_{n+1}$.

    Let $f,e,(x_1,y_1,\ldots)$ be as in the claim and assume $f(n)>t_e(n)$.
    Being an evader, $e$ is a Turing machine, say, the $k$th Turing machine.
    Since $f(n)>t_e(n)$, it follows that $T_k$ is an $n$th-order $e$-lookalike.
    It follows that, on input $(x_1,\ldots,x_n)$, $p_f$ will play the output
    given by some $n$th-order $e$-lookalike $T_{k'}$, $k'\leq k$.

    For any $k'<k$, say that $p_f$ \emph{is tricked by $T_{k'}$ on
    input $(x_1,\ldots,x_n)$} if, on said input, $p$ identifies $T_{k'}$
    as the first $n$th-order $e$-lookalike and so plays $y_{n+1}=X_{k',n}$ but
    $X_{k',n}\not=x_{n+1}$
    (loosely speaking: $p_f$ is led to believe the evader is $T_{k'}$, and
    this false belief causes $p_f$ to incorrectly predict the evader's next digit).
    It follows that $p_f$ will \emph{not} identify $T_{k'}$ as an $n'$th-order
    $e$-lookalike when run on $(x_1,\ldots,x_{n'})$ for any $n'>n$.
    Thus, $p_f$ can only be tricked at most once by $T_{k'}$ for any particular
    $k'<k$. If $p_f$ is not so tricked, then either $p_f$ identifies $T_{k'}$
    as the first $n$th-order $e$-lookalike for some $k'<k$ (in which case
    $p_f$ plays $y_{n+1}=X_{k',n}=x_{n+1}$, lest $p_f$ be tricked by $T_{k'}$), or else
    $p_f$ identifies $T_k$ as the first $n$th-order $e$-lookalike, in which case
    $p_f$ plays $y_{n+1}=X_{k,n}=x_{n+1}$ since $e=T_k$. Either way, after possibly
    finitely many exceptions caused by being tricked, $p_f$ always
    plays $y_{n+1}=x_{n+1}$ whenever
    $f(n)>t_e(n)$, proving the claim.

    We claim $p_{f_m}$ has original Hibbard measure $\geq m$. To see this,
    let $e\in E_{f_m}$, we must show $p$ learns $e$.
    Let $(x_1,y_1,\ldots)$ be the result of $p_{f_m}$ playing against $e$.
    Since $e\in E_{f_m}$, $f_m\succ t_e$, so for-all-but-finitely-many $n\in\mathbb N$,
    $f_m(n)>t_e(n)$. And, by the above Claim, with at most finitely many exceptions,
    whenever $f_m(n)>t_e(n)$, $x_{n+1}=y_{n+1}$. It follows that $p_{f_m}$ learns $e$,
    as desired.
\end{proof}

Unfortunately, the original Hibbard measure is not computable
(unless the background model of computation is contrived), as the following
proposition shows\footnote{In fact, assuming a non-contrived background
model of computation, for any strictly
increasing total computable function $f$,
even the following can be shown to be non-computable:
given an evader $e$, to determine whether or not $f\succ t_e$.}.

\begin{proposition}
\label{originalnoncomputableprop}
    Assume the background model of computation is well-behaved enough that
    there is an evader $e_0$ which always outputs $0$ and whose runtime $t_{e_0}$
    is bounded by some primitive recursive function.
    Then the original Hibbard measure is not computable:
    there is no effectively computable procedure, given a predictor, to compute its
    original Hibbard measure. In fact, there is not even an effectively computable
    procedure to tell if one given predictor has a higher original Hibbard measure
    than another given predictor.
\end{proposition}

\begin{proof}
    Let $p_1$ be a predictor which always outputs $1$, and let $m$ be its
    original Hibbard measure. By the existence of $e_0$, it
    follows that $m<\infty$ since certainly $p_1$ does not learn $e_0$.
    Let $p$ be a predictor with
    original Hibbard measure $>m$
    (Proposition \ref{complicatedproposition}).

    If the proposition were false, then we could solve
    the halting problem as follows.
    Given any Turing machine $M$, in order to determine whether or not $M$ halts,
    proceed as follows. Let $p_M$ be the predictor which, on
    input $(x_1,\ldots,x_n)$, outputs $p(x_1,\ldots,x_n)$
    if $M$ halts in $\leq n$ steps, or $1$ otherwise.
    Clearly, $M$ halts if and only if $p_M$ has a higher original Hibbard
    measure than $p_1$.
\end{proof}

Likewise, the variations on Hibbard's measure which we present in this paper are also
non-computable. To quantify their precise degrees of computability (e.g., where they
fall within the arithmetical hierarchy) would be beyond the scope of this paper.
We will, however, state one conjecture. If we modified the original Hibbard
measure by replacing Liu's enumeration of the primitive recursive functions by an
enumeration of all total computable functions, then we conjecture the resulting measure would
be strictly higher in the arithmetical hierarchy (i.e., would require strictly
stronger oracles to compute), essentially because the set of total computable functions
is \emph{not} computably enumerable, whereas the primitive recursive functions are.

\subsection{Predictor intelligence and AGI intelligence}
\label{agiproxysection}

Definition \ref{classichibbardmeasuredefn}, and similar measures and taxonomies
which we will
define later, quantify the intelligence of predictors in the game of
adversarial sequence prediction.
But any method for quantifying the intelligence of such predictors can also
approximately quantify the intelligence of (suitably idealized)
agents with Artificial General Intellience (that is, the intelligence of AGIs).

The idealized AGIs we have in mind should be capable of understanding,
and obedient in following or trying to follow, commands issued in everyday human
language\footnote{It is somewhat unclear how explicitly an AGI would
obey certain commands. To use an example of
\citet{yampolskiycontrol}, if we asked a car-driving AGI
to stop the car, would the AGI stop the car
in the middle of traffic, or would it pull over to the side first?
We assume this ambiguity does not apply when we ask the AGI to perform
tasks of a sufficiently abstract and mathematical nature.}
(this is not to say that all AGIs must necessarily be obedient, merely that
for the purposes of this paper we restrict our attention to obedient AGIs).
For example, if such an idealized
AGI were commanded, ``until further notice, compute and list the
digits of pi,'' it would be capable of understanding that command, and would
obediently compute said digits until commanded otherwise\footnote{Our thinking
here is reminiscent of some remarks of \citet{yampolskiy2013turing}.}.

It is unclear how an AGI ought to respond if given an impossible command,
such as ``write a computer program
that solves the halting problem'', or Yampolskiy's
``Disobey!'' \citep{yampolskiycontrol}. But an AGI should be capable of
understanding and attempting to obey an open-ended command, provided it is not
impossible. For example, we could command an AGI to ``until further notice,
write an endless poem about trees,'' and the AGI should be able to do so, writing
said poem line-by-line until we tell it to stop. This is despite the fact that the
command is open-ended and under-determined
(there are many decisions involved in writing a
poem about trees, and we have left all these decisions to the AGI's discretion).
The AGI's ability to obey such open-ended and under-determined commands
exemplifies its
ability to ``adapt with insufficient
knowledge and resources'' \citep{wang2019defining}.
One well-known example of an open-ended command which an AGI should be perfectly
capable of attempting to obey (perhaps at peril to us) is
Bostrom's ``manufacture as many paperclips as possible'' \citep{bostrom2003ethical}.

In particular, such an idealized AGI $X$ should be
capable of obeying the following command:
``Act as a predictor in the game of adversarial sequence prediction''.
By giving $X$ this command, and then immediately filtering out all $X$'s
sensory input except only for input about the digits chosen by an evader,
we would obtain a formal predictor in the sense of Definition \ref{evaderpredictordefn}.
This predictor might be called ``the predictor generated by $X$''. Strictly speaking,
if the command is given to $X$ at time $t$, then it would be more proper to call
the resulting predictor ``the predictor generated by $X$ at time $t$'': up until
time $t$, the observations $X$ makes about the universe might have an effect on
the strategy $X$ chooses to take once commanded to act as a predictor; but as long
as we filter $X$'s sensory input immediately after giving $X$ the command, no
further such observations can so alter $X$'s strategy.
In short, to use Yampolskiy's terminology \citep{yampolskiy2012ai}, the act of
trying to predict adversarial sequence evaders is \emph{AI-easy}.

Thus, any intelligence measure (or taxonomy) for predictors also serves as an intelligence
measure (or taxonomy) for suitably idealized
AGIs. Namely: the intelligence level of an AGI $X$ is equal to the
intelligence level of $X$'s predictor. Of course, a priori,
$X$ might be very intelligent at various other things while being poor
at sequence prediction, or vice versa, so this only
approximately captures $X$'s true intelligence.

Of course, the same could be said for any competency measure on any task:
we do not make any claims that
when we measure $X$'s intelligence via $X$'s predictor's performance, that this
is in any sense ``the one true intelligence measure''. One could just as well
measure $X$'s intelligence in terms of the Elo ranking $X$ would obtain if one
ordered $X$ to compete at chess.
We would offer two motivations to consider adversarial sequence prediction
ability as a particularly interesting proxy for AGI intelligence measurement:
\begin{enumerate}
    \item There seem to be high-level connections between intelligence and
    prediction in general \citep{hutter2004universal}, of which adversarial sequence prediction
    is an elegant and parsimonious example.
    \item Adversarial sequence prediction ability is not bounded above, in the sense
    that for any particular predictor $p$, one can easily produce a predictor $p^+$
    that learns all the evaders which $p$ learns and at least one additional
    evader.
\end{enumerate}

\section{Quantifying growth rates of functions}
\label{growthratesection}

The following is a general and open-ended problem.

\begin{problem}
\label{bigoproblem}
    Quantify the growth-rate of functions from $\mathbb N$ to $\mathbb N$.
\end{problem}

The definition of the original Hibbard measure
(Definition \ref{classichibbardmeasuredefn})
can be thought of as implicitly depending on a specific solution to Problem
\ref{bigoproblem}, which we make explicit in the following definition.

\begin{definition}
\label{hibbardgrowthratedefn}
    For each $m>0$, let $f_m$ be as in Definition \ref{classichibbardmeasuredefn}.
    For each $f:\mathbb N\to\mathbb N$, we define the \emph{original Hibbard growth rate}
    $H(f)$ to be $\min\{m>0\,:\,f_m\succ f\}$ if there is any such $m>0$, and otherwise
    $H(f)=\infty$.
\end{definition}

In order to generalize the original Hibbard definition in a uniform way, we will rearrange
notation somewhat. We have, in some sense, more notation than necessary, namely the notation
in \citep{hibbard} and synonymous notation which is modified to generalize more readily.

\begin{lemma}
\label{straightfwdtechnicallemma}
    For every natural $m>0$ and every $f:\mathbb N\to\mathbb N$,
    $H(f)\leq m$ if and only if $f_m\succ f$.
\end{lemma}

\begin{proof}
    Straightforward.
\end{proof}

\begin{definition}
\label{variationondefinitionofEdefn}
    For every $m\in\mathbb N$, let $E^H_m$
    be the set of all evaders $e$ such that $H(t_e)\leq m$.
\end{definition}

\begin{lemma}
\label{equivalenceoftwoevadersetslemma}
    For every natural $m>0$, $E^H_m=E_{f_m}$.
\end{lemma}

\begin{proof}
    Let $e$ be an evader. By Definition \ref{variationondefinitionofEdefn},
    $e\in E^H_m$ if and only if $H(t_e)\leq m$.
    By Lemma \ref{straightfwdtechnicallemma}, $H(t_e)\leq m$ if and only if
    $f_m\succ t_e$. But by Definition \ref{evadersetdefinition}, this is the
    case if and only if $e\in E_{f_m}$.
\end{proof}

\begin{corollary}
\label{rephrasinghibbardsmeasurecorollary}
    For every predictor $p$, the original Hibbard measure of $p$
    is equal to the maximum natural $m>0$ such that
    $p$ learns $e$ whenever $e\in E^H_m$,
    or is equal to $0$ if there is no such $m$,
    or is equal to $\infty$
    if $p$ learns $e$ whenever $e\in E^H_m$ for all $m>0$.
\end{corollary}

\begin{proof}
    Immediate by Lemma \ref{equivalenceoftwoevadersetslemma}
    and Definition \ref{classichibbardmeasuredefn}.
\end{proof}

In other words, if $S$ is the set of all the $m$ as in Corollary
\ref{rephrasinghibbardsmeasurecorollary}, then the original Hibbard
measure of $p$ is the ``canonical upper bound'' of $S$, where
by the ``canonical upper bound'' of a set of natural numbers
we mean the maximum element of that set (or $\infty$ if that set
is unbounded).

\begin{remark}
\label{epiphanyremark}
Corollary \ref{rephrasinghibbardsmeasurecorollary} shows that
the definition of the original Hibbard measure can be rephrased in such a way
as to show that it depends in a uniform way on a particular solution to
Problem \ref{bigoproblem}, namely on the solution proposed by
Definition \ref{hibbardgrowthratedefn}. For \emph{any} solution $H'$ to
Problem \ref{bigoproblem}, we could define evader-sets $E^{H'}_m$ in a similar
way to Definition \ref{variationondefinitionofEdefn}, and, by copying
Corollary \ref{rephrasinghibbardsmeasurecorollary}, we could obtain a corresponding
intelligence measure given by $H'$
(provided there be some way of choosing canonical bounds of bounded sets in the
underlying number system---if not, we would have to be content with a taxonomy
rather than a measure, a predictor's intelligence falling into many nested
taxa corresponding to many different upper bounds, just as in Big-O notation a
function can simultaneously be $O(n^2)$ and $O(n^3)$).
This formalizes what we claimed in the Introduction,
that Hibbard's idea can be decomposed into two sub-ideas, firstly, that a predictor's
intelligence can be classified in terms of the growth rates of the runtimes of the
evaders it learns, and secondly, a particular method
(Definition \ref{hibbardgrowthratedefn})
of measuring those growth rates (i.e., a particular solution to
Problem \ref{bigoproblem}).
\end{remark}

\subsection{A theoretical note on the difficulty of Problem \ref{bigoproblem}}

In this subsection, we will argue that in order for a solution to Problem \ref{bigoproblem}
to be much good, it should probably measure growth rates using some alternative
number system to the real numbers. Essentially,
this is because the real numbers have the Archimedean property
(the property that for any positive real $r>0$ and any real $y$, there is
some $n\in\mathbb N$ such that $nr>y$), a constraint which does
not apply to function growth rates.

\begin{definition}
    Let $\mathbb N^\mathbb N$ be the set of all functions $\mathbb N\to\mathbb N$.
    A \emph{well-behaved real-measure of $\mathbb N^\mathbb N$} is a function
    $F:\mathbb N^\mathbb N\to\mathbb R$ satisfying the following requirements.
    \begin{enumerate}
        \item (Monotonicity)
        For each $f,g:\mathbb N\to\mathbb N$, if $f\succ g$, then $F(f)>F(g)$.
        \item (Nontriviality)
        For each $r\in\mathbb R$, there is some $f:\mathbb N\to\mathbb N$ such that
        $F(f)>r$.
    \end{enumerate}
\end{definition}

\begin{theorem}
\label{nonarchimedeantheorem}
    There is no well-behaved real-measure of $\mathbb N^\mathbb N$.
\end{theorem}

\begin{proof}
    Assume $F:\mathbb N^\mathbb N\to\mathbb R$ is a well-behaved real-measure
    of $\mathbb N^\mathbb N$.
    By Nontriviality, there are
    $f_0,f_1,\ldots:\mathbb N\to\mathbb N$
    such that each $F(f_n)>n$.
    Define $g:\mathbb N\to\mathbb N$ by $g(n)=f_0(n)+\cdots+f_n(n)$.
    Clearly, for every $n\in\mathbb N$, $g\succ f_{n}$.
    By the Archimedean property of the real numbers, there is some $n\in\mathbb N$
    such that $n>F(g)$. By Monotonicity, $F(g)>F(f_{n})$,
    but by choice of $f_{n}$, $F(f_{n})>n>F(g)$, a contradiction.
\end{proof}

In light of Theorem \ref{nonarchimedeantheorem}, we are motivated to investigate
solutions to Problem \ref{bigoproblem} using alternatives to the real numbers,
which will yield ASPI measures (or taxonomies) in terms of those alternatives.

An informal argument
could be made that real numbers might be inadequate for measuring
AGI intelligence in general\footnote{
This argument was pointed out by \citet{ioi1},
and by \citet{alexander2020arch} again, the latter amidst a wider discussion of
Archimedean and non-Archimedean measures.}.
It at least seems plausible that there are AGIs $X_1,X_2,\ldots$
such that each $X_{i+1}$ is significantly more intelligent than $X_i$, and another AGI
$Y$ such that $Y$ is more intelligent than each $X_i$. At least, if this is not true,
it is not \emph{obvious} that it is not true, and it seems like it would be nontrivial
to argue that it is not true\footnote{To do so would require arguing
that $\forall X_1,X_2,\ldots$, if each $X_{i+1}$ is significantly more intelligent
than $X_i$, then $\forall Y$, $\exists i$ such that $Y$ is at most as intelligent
as $X_i$.}. Now, if ``significantly more
intelligent'' implies ``at least $+1$ more intelligent'', then it follows that the
intelligence levels of $Y$ and of $X_1,X_2,\ldots$ could not all be real numbers,
or else one of the $X_i$ would necessarily be more intelligent than $Y$.

If, as the above argument suggests, the real numbers might potentially be
too constrained to perfectly measure intelligence, what next? How could we measure
intelligence other than by real numbers?
A key motivation for the measures and taxonomies we will come up with below is to
provide examples of intelligence measurement using alternative number systems.
It is for this reason that we do not, in this paper, consider variations on Hibbard's
intelligence measure that arise from simply replacing Liu's enumeration of the primitive
recursive functions with various other $\mathbb N$-indexed lists of functions
(for example, the list of all total computable functions).

\section{Big-O and Big-$\Theta$ intelligence}
\label{bigosection}

One of the most standard solutions
to Problem \ref{bigoproblem} in computer science is to categorize
growth rates of arbitrary functions by comparing them to more familiar functions using
Big-O notation or Big-$\Theta$ notation.
\citet{knuth1976big} defines these as follows
(we modify the definition slightly because
we are only concerned here with functions from $\mathbb N$ to $\mathbb N$).

\begin{definition}
\label{bigodefn}
    Suppose $f:\mathbb N\to\mathbb N$. We define the following function-sets.
    \begin{itemize}
        \item
        $O(f(n))$ is the set of all $g:\mathbb N\to\mathbb N$ such that
        there is some real $C>0$ and some $n_0\in\mathbb N$ such that
        for all $n\geq n_0$, $g(n)\leq Cf(n)$.
        \item
        $\Theta(f(n))$ is the set of all $g:\mathbb N\to\mathbb N$ such that
        there are some real $C>0$ and $C'>0$ and some $n_0\in\mathbb N$ such that
        for all $n\geq n_0$, $Cf(n)\leq g(n)\leq C'f(n)$.
    \end{itemize}
\end{definition}

Note that Definition \ref{bigodefn} does not measure growth rates, but rather
categorizes growth rates into Big-O and Big-$\Theta$ taxonomies.
For example, the same function can be
both $O(n^2)$ and $O(n^3)$, the former taxon being nested within the latter.

By Remark \ref{epiphanyremark}, Definition \ref{bigodefn} yields the following elegant
taxonomy of predictor intelligence.

\begin{definition}
\label{bigointelligencedefn}
    Suppose $p$ is a predictor, and suppose $f:\mathbb N\to\mathbb N$.
    \begin{itemize}
        \item
        We say \emph{$p$ has Big-O ASPI measure $O(f(n))$} if
        $p$ learns every evader $e$ such that $t_e$ is $O(f(n))$.
        \item
        We say \emph{$p$ has Big-$\Theta$ ASPI measure $\Theta(f(n))$} if
        $p$ learns every evader $e$ such that $t_e$ is $\Theta(f(n))$.
    \end{itemize}
\end{definition}

\begin{proposition}
    For any computable function $f:\mathbb N\to\mathbb N$, there is a predictor
    $p$ with Big-O ASPI measure $O(f(n))$ and Big-$\Theta$ ASPI measure
    $\Theta(f(n))$.
\end{proposition}

\begin{proof}
    Define $g:\mathbb N\to\mathbb N$ by $g(n)=nf(n)+1$, and let $p_g$ be as in
    the proof of Proposition \ref{complicatedproposition}.
    We claim $p_g$ has Big-O ASPI measure $O(f(n))$. To see this, let $e$ be
    any evader such that $t_e$ is $O(f(n))$. Thus there is some $C\in\mathbb R$ such that
    for all-but-finitely-many $n\in\mathbb N$, $t_e(n)\leq Cf(n)$.
    It follows that $g\succ t_e$. By the Claim in the proof of Proposition
    \ref{complicatedproposition}, for all-but-finitely-many $n\in\mathbb N$,
    if $g(n)>t_e(n)$ then $x_{n+1}=y_{n+1}$, where $(x_1,y_1,\ldots)$ is the
    result of $p_g$ playing against $e$. So in all, with only finitely many
    exceptions, each $x_{n+1}=y_{n+1}$, as desired. The proof that $p_g$ has
    Big-$\Theta$ ASPI measure $\Theta(f(n))$ is similar.
\end{proof}

\section{ASPI measures based on majorization hierarchies}
\label{majorizationsection}

Majorization hierarchies \citep{weiermann2002slow}
provide ordinal-number-valued measures for the growth
rates of certain functions. A majorization hierarchy depends
on many infinite-dimensional parameters. We will describe two
majorization hierarchies up to the ordinal $\epsilon_0$,
using standard choices for the parameters, and the ASPI measures
which they produce.

\begin{definition}
    (Classification of ordinal numbers)
    Ordinal numbers are divided into three types:
    \begin{enumerate}
        \item Zero: The ordinal $0$.
        \item Successor ordinals: Ordinals of the form $\alpha+1$ for some ordinal $\alpha$.
        \item Limit ordinals: Ordinals which are not successor ordinals nor $0$.
    \end{enumerate}
\end{definition}

For example, the smallest infinite ordinal, $\omega$, is a limit ordinal. It is not zero
(because zero is finite),
nor can it be a successor ordinal, because if it were a successor ordinal, say, $\alpha+1$,
then $\alpha$ would be finite (since $\omega$ is the \emph{smallest} infinite ordinal),
but then $\alpha+1$ would be finite as well.

Ordinal numbers have an arithmetical structure: two ordinals $\alpha$ and $\beta$
have a sum $\alpha+\beta$, a product $\alpha\cdot \beta$, and a power
$\alpha^\beta$. It would be beyond the scope of this paper to give the full
definition of these operations. We will only remark that some care is needed because
although ordinal arithmetic is associative---e.g.,
$(\alpha+\beta)+\gamma=\alpha+(\beta+\gamma)$, and similarly for multiplication---it is
not generally commutative: $\alpha+\beta$ is not always equal to $\beta+\alpha$,
and $\alpha\cdot\beta$ is not always equal to $\beta\cdot\alpha$. For this reason,
one often sees products like $\alpha\cdot 2$, which are not necessarily equivalent to the
more familiar $2\cdot\alpha$.

The ordinal $\epsilon_0$ is the smallest ordinal bigger than the ordinals
$\omega,\omega^\omega,\omega^{\omega^\omega},\ldots$. It satisfies the equation
$\epsilon_0=\omega^{\epsilon_0}$ and can be intuitively thought of as
\[
    \epsilon_0 = \omega^{\omega^{\omega^{\iddots}}}.
\]
Ordinals below $\epsilon_0$ include such ordinals as $\omega$,
$\omega^{\omega+1}+\omega^{\omega}+\omega^5+3$,
\[
\omega^{\omega^{\omega^{\omega^\omega}}}+
\omega^{\omega^{\omega^\omega}+\omega^{\omega\cdot 2+1}+\omega^4 + 3}
+ \omega^{\omega^5+\omega^3}+\omega^8+1,
\]
and so on.
Any ordinal below $\epsilon_0$ can be uniquely written in the form
\[
    \omega^{\alpha_1}+\omega^{\alpha_2}+\cdots + \omega^{\alpha_k}
\]
where $\alpha_1\geq\cdots\geq\alpha_k$ are smaller ordinals below $\epsilon_0$---this form
for an ordinal below $\epsilon_0$ is called its \emph{Cantor normal form}.
For example, the Cantor normal form for $\omega^{\omega\cdot 2}\cdot 2+\omega\cdot 3+2$
is
\[
\omega^{\omega\cdot 2}\cdot 2+\omega\cdot 3+2
=
\omega^{\omega\cdot 2} + \omega^{\omega\cdot 2} + \omega^1 + \omega^1 + \omega^1
+\omega^0 + \omega^0.
\]

\begin{definition}
\label{fundsequencesdefn}
    (Standard fundamental sequences for limit ordinals $\leq\epsilon_0$)
    Suppose $\lambda$ is a limit ordinal $\leq\epsilon_0$. We define a
    \emph{fundamental sequence for $\lambda$},
    written $(\lambda[0],\lambda[1],\lambda[2],\ldots)$, inductively as follows.
    \begin{itemize}
        \item
        If $\lambda=\epsilon_0$, then $\lambda[0]=\omega$,
        $\lambda[1]=\omega^\omega$, $\lambda[2]=\omega^{\omega^\omega}$,
        and so on.
        \item
        If $\lambda$ has Cantor normal form
        $\omega^{\alpha_1}+\cdots+\omega^{\alpha_k}$ where $k>1$,
        then
        each
        \[
            \lambda[i] = \omega^{\alpha_1}+\cdots+\omega^{\alpha_{k-1}}
            + (\omega^{\alpha_k}[i]).
        \]
        \item
        If $\lambda$ has Cantor normal form $\omega^{\alpha+1}$,
        then each $\lambda[i]=\omega^{\alpha}\cdot i$.
        \item
        If $\lambda$ has Cantor normal form $\omega^{\lambda_0}$ where $\lambda_0$
        is a limit ordinal, then each $\lambda[i]=\omega^{\lambda_0[i]}$.
    \end{itemize}
\end{definition}

\begin{example}
    (Fundamental sequence examples)
    \begin{itemize}
        \item
        The fundamental sequence for $\lambda=\omega=\omega^{0+1}$ is
        $\omega^0\cdot 0, \omega^0\cdot 1, \omega^0\cdot 2, \ldots$,
        i.e., $0, 1, 2, \ldots$.
        \item
        The fundamental sequence for $\lambda=\omega^5$ is
        $0,\omega^4,\omega^4\cdot 2,\omega^4\cdot 3,\ldots$.
        \item
        The fundamental sequence for $\lambda=\omega^\omega$ is
        $\omega^0,\omega^1,\omega^2,\ldots$.
        \item
        The fundamental sequence for $\lambda=\omega^\omega+\omega$ is
        $\omega^\omega+0,\omega^\omega+1,\omega^\omega+2,\ldots$.
    \end{itemize}
\end{example}

\begin{definition}
\label{slowgrowinghierarchydefn}
    (The standard slow-growing hierarchy up to $\epsilon_0$)
    We define functions $g_\beta:\mathbb N\to\mathbb N$ (for all ordinals
    $\beta\leq \epsilon_0$) by transfinite induction as follows.
    \begin{itemize}
        \item
        $g_0(n)=0$.
        \item
        $g_{\alpha+1}(n) = g_\alpha(n) + 1$ if $\alpha+1\leq\epsilon_0$.
        \item
        $g_{\lambda}(n) = g_{\lambda[n]}(n)$ if $\lambda\leq\epsilon_0$ is a limit ordinal.
    \end{itemize}
\end{definition}

Here are some early levels in the slow-growing hierarchy, spelled out in detail.

\begin{example}
\label{highdetailslowgrowingexample}
    (Early examples of functions in the slow-growing hierarchy)
    \begin{enumerate}
        \item
        $g_1(n)=g_{0+1}(n)=g_0(n)+1=0+1=1$.
        \item
        $g_2(n)=g_{1+1}(n)=g_1(n)+1=1+1=2$.
        \item
        More generally, for all $m\in\mathbb N$,
        $g_m(n)=m$.
        \item
        $g_\omega(n)=g_{\omega[n]}(n)=g_n(n)=n$.
        \item
        $g_{\omega+1}(n)=g_{\omega}(n)+1=n+1$.
        \item
        More generally, for all $m\in\mathbb N$,
        $g_{\omega+m}(n)=n+m$.
        \item
        $g_{\omega\cdot 2}(n)=g_{(\omega\cdot 2)[n]}(n)
        =g_{\omega+n}(n)=n+n=n\cdot 2$.
    \end{enumerate}
\end{example}

Following Example \ref{highdetailslowgrowingexample}, the reader should be able
to fill in the details in the following example.

\begin{example}
    (More examples from the slow-growing hierarchy)
    \begin{enumerate}
        \item
        $g_{\omega^2}(n)=n^2$.
        \item
        $g_{\omega^3}(n)=n^3$.
        \item
        $g_{\omega^\omega}(n)=n^n$.
        \item
        $g_{\omega^{\omega\cdot 3+1}+\omega+5}(n)=n^{3n+1}+n+5$.
        \item
        $g_{\omega^{\omega^{\omega}}}(n)=n^{n^n}$.
    \end{enumerate}
\end{example}

What about $g_{\epsilon_0}$? Thinking of $\epsilon_0$ as
\[\omega^{\omega^{\omega^{\iddots}}},\]
one might expect $g_{\epsilon_0}(n)$ to be
\[n^{n^{n^{\iddots}}},\]
but such an infinite tower
of natural number exponents makes no sense if $n>1$. Instead,
the answer defies familiar mathematical
notation.

\begin{example}
\label{epsilon0example}
(Level $\epsilon_0$ in the slow-growing hierarchy)
The values of $g_{\epsilon_0}$ are as follows:
\begin{itemize}
    \item
    $g_{\epsilon_0}(0)=0$.
    \item
    $g_{\epsilon_0}(1)=1^1$.
    \item
    $g_{\epsilon_0}(2)=2^{2^2}$.
    \item
    $g_{\epsilon_0}(3)=3^{3^{3^3}}$.
    \item
    And so on.
\end{itemize}
\end{example}

Examples \ref{highdetailslowgrowingexample}--\ref{epsilon0example} illustrate
how the slow-growing hierarchy systematically provides a family of reference
functions against which any particular function can be compared.
This yields a solution to Problem \ref{bigoproblem}: we can declare the growth
rate of an arbitrary function $f:\mathbb N\to\mathbb N$ to be the smallest ordinal
$\beta< \epsilon_0$ such that $g_\beta\succ f$ (or $\infty$ if there is no such
$\beta$).
For any bounded set $S$ of ordinals, there is a canonical upper bound for $S$,
namely, the supremum of $S$.
Thus we obtain an ASPI measure (not just a taxonomy).

\begin{definition}
\label{tradmajorizationhierarchyhibbardmeasuredefn}
    If $p$ is a predictor, the \emph{ASPI measure of $p$ given by the
    standard slow-growing hierarchy up to $\epsilon_0$} is defined to be the
    supremum of $S$ (or $\infty$
    if $\epsilon_0\in S$), where $S$ is the set of all ordinals
    $\alpha\leq\epsilon_0$
    such that the following condition holds:
    for every evader $e$, if $g_\alpha\succ t_e$, then $p$ learns $e$.
\end{definition}

In Definition \ref{slowgrowinghierarchydefn}, in the successor ordinal case,
we chose to define $g_{\alpha+1}(n)=g_\alpha(n)+1$. The resulting majorization
hierarchy is referred to as \emph{slow-growing} because in some sense this
makes $g_{\alpha+1}$ just barely faster-growing than $g_\alpha$.
Different definitions of $g_{\alpha+1}$ would yield different majorization
hierarchies, such as the following.

\begin{definition}
\label{fastgrowinghierarchydefn}
    (The standard fast-growing hierarchy up to $\epsilon_0$, also known as
    the Wainer hierarchy)
    We define functions $h_\beta:\mathbb N\to\mathbb N$ (for all ordinals
    $\beta\leq \epsilon_0$) by transfinite induction as follows.
    \begin{itemize}
        \item
        $h_0(n)=n+1$.
        \item
        $h_{\alpha+1}(n) = h^n_\alpha(n)$, where $h^n_\alpha$ is the $n$th
        iterate of $h_\alpha$ (so $h^1_\alpha(x)=h_\alpha(x)$,
        $h^2_\alpha(x)=h_\alpha(h_\alpha(x))$,
        $h^3_\alpha(x)=h_\alpha(h_\alpha(h_\alpha(x)))$, and so on).
        \item
        $h_{\lambda}(n) = h_{\lambda[n]}(n)$ if $\lambda$ is a
        limit ordinal $\leq\epsilon_0$.
    \end{itemize}
\end{definition}

The functions in the fast-growing hierarchy grow quickly
as $\alpha$ grows. It can be shown \citep{wainer1987provably} that
for every computable function $f$ whose totality can be proven from the axioms of
Peano arithmetic, there is some $\alpha<\epsilon_0$ such that $h_\alpha\succ f$.

\begin{definition}
\label{fastmajorizationhierarchyhibbardmeasuredefn}
    If $p$ is a predictor, the \emph{ASPI measure of $p$ given by the
    standard fast-growing hierarchy up to $\epsilon_0$} is defined to be the
    supremum of $S$ (or $\infty$ if $\epsilon_0\in S$),
    where $S$ is the set of all ordinals $\alpha\leq\epsilon_0$ such that
    the following condition holds:
    for every predictor $e$, if $h_\alpha\succ t_e$, then $p$ learns $e$.
\end{definition}

\begin{proposition}
    For each $\alpha<\epsilon_0$, there is a predictor $p$ (resp.\ $q$)
    whose ASPI measure
    given by the standard slow-growing (resp.\ fast-growing)
    hierarchy up to $\epsilon_0$
    is $\geq \alpha$.
\end{proposition}

\begin{proof}
    Similar to the proof of Proposition \ref{complicatedproposition}.
\end{proof}

Between Definitions \ref{tradmajorizationhierarchyhibbardmeasuredefn} and
\ref{fastmajorizationhierarchyhibbardmeasuredefn}, the former offers a finer
granularity
intelligence measure for the predictors to which it assigns non-$\infty$
intelligence, but the latter assigns non-$\infty$ intelligence to
more intelligent predictors.

Definitions \ref{slowgrowinghierarchydefn}
and \ref{fastgrowinghierarchydefn} are only two
examples of
majorization hierarchies. Both the slow- and fast-growing hierarchies can be
extended by extending the fundamental sequences of Definition
\ref{fundsequencesdefn} to larger ordinals\footnote{Remarkably,
the slow-growing hierarchy eventually catches up with the fast-growing hierarchy
if both hierarchies are extended to sufficiently large ordinals
\citep{wainer1989slow, girard1981pi12}, a beautiful illustration of
how counter-intuitive
large ordinal numbers can be.}, however, the larger
the ordinals become, the more difficult it is to do this, and especially the less
clear it is how to do it in any sort of canonical way.
There are also other choices for how to proceed at successor ordinal stages besides
$g_{\alpha+1}(n)=g_\alpha(n)+1$ or $h_{\alpha+1}(n)=h^n_\alpha(n)$---for example,
one of the oldest majorization hierarchies is the Hardy hierarchy
\citep{hardy1904theorem}, where $H_{\alpha+1}(n)=H_\alpha(n+1)$.
And even for ordinals up to $\epsilon_0$,
there are other ways to choose fundamental sequences besides how we defined them in
Definition \ref{fundsequencesdefn}---choosing non-canonical fundamental sequences can
drastically alter the resulting majorization hierarchy \citep{weiermann1997sometimes}.
All these different majorization hierarchies yield different ASPI measures.

\subsection{A remark about ASPI measures and AGI intelligence}

All the ASPI measures and taxonomies we have defined so far double as indirect
intelligence measures and taxonomies for an AGI, by the argument we made
in Subsection \ref{agiproxysection}.

For a given AGI $X$, a priori, we cannot say much about the predictor which $X$
would act as if $X$ were commanded to act as a predictor. But there is one particularly
elegant and parsimonious strategy which $X$ might use, a \emph{brute force strategy},
namely:
\begin{itemize}
    \item Enumerate all the
    computable functions $f$ which $X$ knows to be total, and for each one, attempt to
    predict the evader $e$ by assuming that the evader's runtime $t_e$
    satisfies $f\succ t_e$.
    If the evader proves not to be so majorized (by differing from every computable
    function whose runtime is so majorized), then move on to the next
    known total function $f$, and continue the process.
\end{itemize}
We do not
know for certain which predictor $X$ would imitate when commanded to act as a predictor,
but it seems plausible that $X$ would use this brute force strategy or something
equivalent.

For an AGI $X$ who uses the above brute force strategy, ASPI
measures of $X$'s intelligence would be determined by $X$'s knowledge, namely,
by the runtime complexity of the computable functions which $X$ knows to be total.
Furthermore, the most natural way for $X$ to know totality of functions with large
runtime complexity, is for $X$ to know fundamental sequences for large ordinal
numbers, and produce said functions by means of majorization
hierarchies\footnote{It may be
possible for an AGI to be contrived to know totality of functions that are larger
than the functions produced by majorization hierarchies up to ordinals the AGI knows
about, but we conjecture that that is not the case for AGIs not so deliberately
contrived.}. This suggests a connection between
\begin{enumerate}
    \item
    ASPI measures like that of
    Definition \ref{fastmajorizationhierarchyhibbardmeasuredefn}, and
    \item
    intelligence measures based on which ordinals the AGI knows
    \citep{ioi1}.
\end{enumerate}
Indeed, \citet{ioi2} has argued that
the task of notating large ordinals is one which
spans the entire range of intelligence.
This is reminiscent of Chaitin's proposal to use ordinal notation
as a goal intended to facilitate evolution---``and the larger the ordinal,
the fitter the organism'' \citep{chaitin}---and Good's observation
\citep{good1969godel} that iterated Lucas-Penrose contests boil down to
contests to name the larger ordinal.

\section{Hyperreal numbers and surreal numbers}
\label{hyperrealsurrealsection}

In this section, we will exhibit an abstract ASPI taxonomy based on hyperreal numbers
and an abstract ASPI measure based on surreal numbers. We do not assume previous
familiarity with either of these number systems.

\subsection{The hyperreal ASPI taxonomy}
\label{hyperrealsubsection}

In this subsection, we will begin by considering growth rate \emph{comparison}, which is
a strictly simpler problem than growth rate \emph{measurement} (our proposed solution will
then lead to a numerical growth rate measure anyway). Given two functions $f$
and $g$, does $f$ outgrow $g$ or does $f$ not outgrow $g$? We would like to say that
$f$ outgrows $g$ if and only if $f(n)>g(n)$ for ``a majority of'' $n\in\mathbb N$, but it is
not clear what ``majority'' should mean. Certainly if $f(n)>g(n)$ for all but finitely many
$n\in\mathbb N$, it should be safe to say $f$ outgrows $g$, and if $f(n)\leq g(n)$ for
all but finitely many $n\in\mathbb N$, it should be safe to say $f$ does not outgrow
$g$. But what if there are infinitely many $n\in\mathbb N$ such that $f(n)>g(n)$, and
infinitely many $n\in\mathbb N$ such that $f(n)\leq g(n)$?

Adapting the key insight from \citet{alexander2019intelligence}, consider each $n\in\mathbb N$
to be a voter in an election to determine whether or not $f$ outgrows $g$.
Each $n$ votes based on whether or not $f(n)>g(n)$. For example, $532$ is a voter in this
election. If $f(532)>g(532)$, then $532$ casts her vote for ``$f$ outgrows $g$''; otherwise,
$532$ casts her vote for ``$f$ does not outgrow $g$''. This reduces the outgrowth problem
to an election decision problem: $f$ shall be considered to outgrow $g$ if and only if
``$f$ outgrows $g$'' gets a winning bloc of votes.

We need to decide what it means for a set $N\subseteq \mathbb N$ to constitute a winning
bloc of votes. We reason as follows.
\begin{itemize}
    \item $\emptyset$ should not be a winning bloc: if no-one votes for
    you, you lose.
    \item If $N_1$ is a winning bloc and $N_1\subseteq N_2$, then $N_2$ should
    be a winning bloc: if you were already winning, and additional voters switch
    their votes to you, you should still win.
    \item For any $N\subseteq \mathbb N$, either $N$ should be a winning bloc,
    or its complement $N^c=\mathbb N\backslash N$ should be a winning bloc:
    however the election goes, either you win or your opponent wins.
    \item We should insist on the outgrowth relation being transitive:
    if $f$ outgrows $g$ and $g$ outgrows $h$, then $f$ should outgrow $h$.
    Suppose that
    \begin{align*}
        N_{fg} &= \{ n\in\mathbb N \,:\, f(n)>g(n)\},\\
        N_{gh} &= \{ n\in\mathbb N \,:\, g(n)>h(n)\},\mbox{ and}\\
        N_{fh} &= \{ n\in\mathbb N \,:\, f(n)>h(n)\}.
    \end{align*}
    Clearly $N_{fg}\cap N_{gh}\subseteq N_{fh}$ but, a priori, we cannot say
    more: one can find functions $f,g,h$ such that
    $N_{fg}\cap N_{gh}=N_{fh}$. Thus, in order to ensure transitivity of the
    outgrowth relation, we should insist on the following requirement.
    Whenever $N_1$ and $N_2$ are winning blocs, then $N_1\cap N_2$ should be a winning bloc.
    \item We could trivially satisfy the
    above requirements, namely: we could choose some $n_0\in\mathbb N$ and
    declare that $N\subseteq \mathbb N$ is a winning bloc if and only if
    $n_0\in N$. In electoral terms, this would amount to making $n_0$ a dictator, whose
    vote decides the election regardless how anyone else votes. In terms of
    the outgrowth relation, this would amount to declaring that $f$ outgrows $g$
    if and only if $f(n_0)>g(n_0)$. That would be a poor method
    of comparing growth rates. Thus, we should insist that $\{n_0\}$ is not a winning
    bloc for any $n_0\in\mathbb N$.
\end{itemize}
Is it possible to satisfy all the above requirements, or are they too demanding?
It turns out it is possible. In fact, the above requirements are exactly the requirements
of a \emph{free ultrafilter}, an important device from mathematical logic.

\begin{definition}
\label{ultrafilterdefn}
    An \emph{ultrafilter on $\mathbb N$} (or more simply an \emph{ultrafilter})
    is a set $\mathcal U$ of subsets of $\mathbb N$ such that:
    \begin{enumerate}
        \item $\emptyset\not\in\mathcal U$.
        \item For every $N_1\in\mathcal U$, for every $N_2\subseteq \mathbb N$,
        if $N_1\subseteq N_2$, then $N_2\in\mathcal U$.
        \item For every $N\subseteq\mathbb N$, either $N\in\mathcal U$
        or $N^c\in\mathcal U$.
        \item ($\cap$-closure) For every $N_1,N_2\in\mathcal U$, $N_1\cap N_2\in \mathcal U$.
    \end{enumerate}
    An ultrafilter is \emph{free} if it does not contain any singleton
    $\{n_0\}$ ($n_0\in\mathbb N$).
\end{definition}

Clearly a free ultrafilter is exactly a notion of winning blocs meeting all our
requirements.
The following theorem is well-known in mathematical logic, and we state it without
proof.

\begin{theorem}
\label{freeultrafiltersexistthm}
    Free ultrafilters exist.
\end{theorem}

Theorem \ref{freeultrafiltersexistthm} is profound because
it is counter-intuitive that there should be a non-dictatorial method of determining
election winners satisfying $\cap$-closure. To see how counter-intuitive $\cap$-closure
is, suppose that in 2021 the Dog party wins the presidency and in 2022 the
Cat party wins the presidency (with the same voters every year and no other parties).
Call a voter a ``Dog-to-Cat switcher'' if they vote Dog in 2021 and Cat in 2022.
The $\cap$-closure property says in order to win in 2023, it would be enough to
get just the Dog-to-Cat switchers' votes \emph{and no others}.
For more on infinite-voter elections and free ultrafilters,
and especially their interplay with Arrow's impossibility theorem,
see \citet{kirman}.

For the remainder of the section, let $\mathcal U$ be a free ultrafilter.
Unfortunately, logicians have shown that, though free ultrafilters exist,
it is impossible to concretely exhibit one. More precisely, all known proofs
of Theorem \ref{freeultrafiltersexistthm} are non-constructive (using non-constructive
set-theoretic axioms such as the Axiom of Choice) and logicians have proven that
Theorem \ref{freeultrafiltersexistthm} cannot be proved constructively.

\begin{definition}
\label{outgrowsdefn}
    Suppose $f,g:\mathbb N\to\mathbb R$.
    We say $f>_\mathcal U g$ if
    \[\{n\in\mathbb N\,:\,f(n)>g(n)\}\in \mathcal U.\]
    In other words: if $\mathcal U$ is thought of as a black box deciding which
    subsets of $\mathbb N$ are winning blocs, then $f>_\mathcal U g$
    if and only if ``$f$ outgrows $g$'' wins the election when
    each $n\in\mathbb N$ votes for ``$f$ outgrows $g$'' or ``$f$ does not outgrow $g$''
    depending whether $f(n)>g(n)$ or $f(n)\leq g(n)$ respectively.
\end{definition}

\begin{lemma}
\label{transitivitylemma}
    $>_\mathcal U$ is transitive.
\end{lemma}

\begin{proof}
    Suppose $f,g,h:\mathbb N\to\mathbb R$
    are such that $f>_\mathcal U g$ and $g>_\mathcal U h$, we must show
    $f>_\mathcal U h$.
    Let
    \begin{align*}
        N_{fg} &= \{n\in\mathbb N\,:\,f(n)>g(n)\},\\
        N_{gh} &= \{n\in\mathbb N\,:\,g(n)>h(n)\},\mbox{ and}\\
        N_{fh} &= \{n\in\mathbb N\,:\,f(n)>h(n)\}.\\
    \end{align*}
    Since $f>_\mathcal U g$, $N_{fg}\in\mathcal U$.
    Since $g>_\mathcal U h$, $N_{gh}\in\mathcal U$.
    By $\cap$-closure, $N_{fg}\cap N_{gh}\in\mathcal U$.
    Clearly $N_{fg}\cap N_{gh}\subseteq N_{fh}$, so, by (2) of
    Definition \ref{ultrafilterdefn},
    $N_{fh}\in \mathcal U$, that is, $f>_\mathcal U h$.
\end{proof}

We will now explain how Definition \ref{outgrowsdefn} leads to a numerical
growth rate measure and, in turn, an ASPI taxonomy.
We will show that by coming up with Definition \ref{outgrowsdefn}, we have actually
done much of the work of the so-called ultrapower construction of
the hyperreal number system, studied in the field of non-standard
analysis \citep{robinson, goldblatt2012lectures}.

\begin{definition}
\label{equivrelndefn}
(Compare Definition \ref{outgrowsdefn})
    Suppose $f,g:\mathbb N\to\mathbb R$. We say
    $f\equiv g$ if
    \[
    \{n\in\mathbb N\,:\, f(n) = g(n)\} \in \mathcal U.
    \]
    In other words, $f\equiv g$ if ``$f=g$'' wins the election (as decided by
    $\mathcal U$) when each
    $n\in\mathbb N$ votes for ``$f=g$'' or ``$f\not=g$'' depending whether
    $f(n)=g(n)$ or $f(n)\neq g(n)$ respectively.
\end{definition}

\begin{lemma}
    The relation $\equiv$ (from Definition \ref{equivrelndefn}) is an equivalence
    relation.
\end{lemma}

\begin{proof}
    Symmetry and reflexivity are trivial.
    The proof of transitivity is similar to Lemma \ref{transitivitylemma}.
\end{proof}

\begin{definition}
\label{hyperrealsdefn}
    The \emph{hyperreal numbers}, written ${}^*\mathbb R$,
    are the equivalence classes of $\equiv$.
    For every $f:\mathbb N\to\mathbb R$, write $\hat f$ for
    the hyperreal number (i.e., the $\equiv$-equivalence class) containing $f$.
    We endow ${}^*\mathbb R$ with arithmetic and order as follows
    (where $f,g:\mathbb N\to\mathbb R$):
    \begin{itemize}
        \item We define addition on ${}^*\mathbb R$ by declaring that
        $\hat {f}+\hat {g}=\hat {h}$
        where $h(n)=f(n)+g(n)$.
        \item We define multiplication on ${}^*\mathbb R$ by declaring that
        $\hat {f}\cdot \hat {g}=\hat {h}$
        where $h(n)=f(n)g(n)$.
        \item We order ${}^*\mathbb R$ by declaring that
        $\hat {f}>\hat {g}$ if and only if $f>_\mathcal U g$ (Definition \ref{outgrowsdefn}).
    \end{itemize}
\end{definition}

The following is well-known and we state it
without proof.

\begin{theorem}
    The addition, multiplication, and ordering in Definition \ref{hyperrealsdefn}
    are well-defined, and they make ${}^*\mathbb R$ an ordered field.
\end{theorem}

With this machinery,
we now have a trivial hyperreal solution to Problem \ref{bigoproblem}.

\begin{definition}
\label{hyperrealgrowthratedefn}
    (The hyperreal solution to Problem \ref{bigoproblem})
    For any function $f:\mathbb N\to\mathbb N$, the \emph{hyperreal growth rate
    of $f$} is the hyperreal number $\hat f$.
\end{definition}

Because of the non-constructive nature of free ultrafilters, the following notions
are even less practical than the measures in the previous sections.
However, they could potentially be useful for
proving theoretical properties about the intelligence of predictors.

\begin{definition}
\label{Ulearndefn}
    Suppose $p$ is a predictor and $e$ is an evader. Let
    $(x_1,y_1,x_2,y_2,\ldots)$ be the result of $p$ playing against $e$.
    We say $p$ \emph{$\mathcal U$-learns} $e$ if
    \[\{n\in\mathbb N\,:\,p(x_1,\ldots,x_n)=e(y_1,\ldots,y_n)\}\in\mathcal U\]
    (or equivalently: $\{n\in\mathbb N\,:\,y_{n+1}=x_{n+1}\}\in\mathcal U$).
    In other words, $p$ $\mathcal U$-learns $e$ if ``$p$ learns $e$''
    wins the election (according to $\mathcal U$)
    when every $n\in\mathbb N$ votes for ``$p$ learns $e$''
    or ``$p$ does not learn $e$'' depending whether or not
    $p(x_1,\ldots,x_n)=e(y_1,\ldots,y_n)$.
\end{definition}

In the following
definition, rather than assigning a particular hyperreal number intelligence to every
predictor, rather, we categorize predictors into a taxonomy.
This is necessary because there is no way of choosing canonical bounds
of bounded sets of hyperreal numbers in general. For lack of a way of
choosing a particular bound, we are forced to consider many taxa corresponding
to many bounds.

\begin{definition}
\label{hyperrealhibbardintelligencedefn}
    (The hyperreal ASPI taxonomy)
    Let $p$ be a predictor and let $\hat f$ be a hyperreal number.
    We say that $p$ \emph{has hyperreal ASPI intelligence at least $\hat f$}
    if and only if the following condition holds:
    for every evader $e$, if the hyperreal growth rate of $t_e$ is
    $<\hat f$, then $p$ $\mathcal U$-learns $e$.
\end{definition}

Now we would like to state an analog of Proposition \ref{complicatedproposition}
for hyperreal ASPI intelligence, but before we can do that, we need to state the
following lemma. This lemma is well-known so we state it without proof.

\begin{lemma}
\label{technicallemmaaboutultrafilters}
    For any $N\in\mathcal U$, for any finite $N_0\subseteq\mathbb N$,
    the difference $N\backslash N_0$ is in $\mathcal U$.
\end{lemma}

\begin{proposition}
    For any computable function $f:\mathbb N\to\mathbb N$, there is a predictor which has
    hyperreal ASPI intelligence at least $\hat f$.
\end{proposition}

\begin{proof}
    Let $p_f$ be as in the proof of Proposition \ref{complicatedproposition}.
    We claim $p_f$ has hyperreal ASPI intelligence at least $\hat f$.
    To see this, assume $e$ is an evader such that the hyperreal growth rate of
    $t_e$ is $<\hat f$, we will show $p_f$ $\mathcal U$-learns $e$.

    Let $(x_1,y_1,x_2,y_2,\ldots)$ be the result of $p_f$ playing against $e$.
    Let
    \begin{align*}
        N_1 &= \{n\in\mathbb N\,:\,f(n)>t_e(n)\},\\
        N_2 &= \{n\in N_1\,:\,p_f(x_1,\ldots,x_n)=e(y_1,\ldots,y_n)\},\\
        N_3 &= \{n\in\mathbb N\,:\,p_f(x_1,\ldots,x_n)=e(y_1,\ldots,y_n)\}.
    \end{align*}
    By the Claim in the proof of Proposition \ref{complicatedproposition},
    for all-but-finitely-many $n\in\mathbb N$, if $f(n)>t_e(n)$, then
    $x_{n+1}=y_{n+1}$, i.e., $p_f(x_1,\ldots,x_n)=e(y_1,\ldots,y_n)$.
    In other words, $N_2=N_1\backslash N_0$ for some finite $N_0\subseteq \mathbb N$.

    Since the hyperreal growth rate of $t_e$ is $<\hat f$, $N_1\in\mathcal U$.
    Since $N_2=N_1\backslash N_0$, by Lemma \ref{technicallemmaaboutultrafilters}
    we have $N_2\in\mathcal U$. Since $N_2\subseteq N_3$, it follows that
    $N_3\in\mathcal U$, that is, $p_f$ $\mathcal U$-learns $e$, as desired.
\end{proof}

Whereas we showed in Proposition \ref{originalnoncomputableprop} that
the original Hibbard measure is already
non-computable, we conjecture that the hyperreal ASPI taxonomy is even harder
(in a computability theoretical sense).

We mention in passing that there are other ways to approach non-standard analysis
\citep{katz},
where the free ultrafilter dependency is replaced by a dependency on a model of
a certain set of axioms.

\subsection{The surreal ASPI measure}

In Definition \ref{hyperrealhibbardintelligencedefn}, we had to content ourselves
with an ASPI taxonomy rather than an ASPI measure, because there is no canonical
way of choosing a preferred bound for a bounded set of hyperreals.
In this section, we will consider another number system, the surreal numbers
\citep{conway, knuth, ehrlich2012absolute}, in which the hyperreal numbers can be
embedded. By embedding the hyperreals within the surreals, our hyperreal solution
to Problem \ref{bigoproblem} yields a surreal solution to Problem \ref{bigoproblem},
via the embedding.
Unlike the hyperreal numbers, the surreal numbers \emph{do} admit a
canonical way of choosing preferred bounds for bounded sets. Thus, our surreal
solution to Problem \ref{bigoproblem} will yield an ASPI measure, not just a taxonomy.

Formally defining the surreal numbers involves nuances touching the foundations
of mathematics, so we will only sketch the definition here.
Suppose we take the following as the guiding principle in creating a number system:
\begin{itemize}
    \item
    For every set $L$ of lower bounds in our number system, and every set $R$ of
    upper bounds in our number system, with $L<R$ (by which we mean $\ell<r$
    for all $\ell\in L, r\in R$), we want there to exist a canonical number
    $L<x<R$ (i.e., a canonical number $x$ such that $\ell<x<r$ for all
    $\ell\in L, r\in R$).
\end{itemize}
To simplify things, since this guiding principle only requires the existence of
these canonical numbers, we can assume no other numbers exist, only such canonical
numbers. Thus, every number in the resulting system can, recursively, be
\emph{identified} with a pair $(L,R)$ of sets of numbers, $L<R$. Notationally,
we use $\{L|R\}$ as a name for the canonical number $L<\{L|R\}<R$ between $L$ and $R$,
whenever $L<R$ are sets of numbers. There are two tricky issues with this idea:
\begin{enumerate}
    \item
    A number in this number system we are building might have multiple names,
    so we cannot simply take the numbers to be their names. Instead, it is necessary
    to take the numbers to be equivalence classes of names.
    What should it mean for two names to be equivalent?
    \item
    If $\{x_L|x_R\}$ and $\{y_L|y_R\}$ are (names of) two numbers $x,y$ (respectively)
    in this number system we are building
    (so $x_L<x_R$ and $y_L<y_R$ are sets of numbers in said number system),
    what does it mean to say $x<y$? This question seems
    somewhat circular, because in order for $\{x_L|x_R\}$ to be a valid name in the
    first place already requires that $x_L<x_R$ (i.e., that $\ell<r$ whenever
    $\ell\in x_L,r\in x_R$).
\end{enumerate}
To answer issue 2, we ask ourselves: what does it mean for the canonical number
$x$ between $x_L$ and $x_R$ to be less than the canonical number $y$ between $y_L$ and $y_R$?
With some thought, it seems a natural way to (recursively)
answer this question is to declare $x<y$ if
and only if one of the following holds:
\begin{itemize}
    \item There is some $y'\in y_L$ such that $x\leq y'$, or
    \item There is some $x'\in x_R$ such that $x'\leq y$.
\end{itemize}
Having answered (2), we can answer (1) by declaring that two names
$\{x_L|x_R\}$ and $\{y_L|y_R\}$ (of $x$ and $y$ respectively) are equivalent
if $x\not<y$ and $y\not< x$.

Carrying out the above definition in full formality is tricky,
because the $<$ relation and the equivalence relation have to be
defined by simultaneous recursion in terms of each other.
To give the formal definition here would be beyond the scope of this
paper\footnote{Technically, $<$ and the equivalence relation are not actually
relations at all, because their
universes are not sets. For this and other reasons, the formal construction of the surreals
is usually carried out in transfinitely many stages, such that at any particular stage,
the surreals constructed so far form a set.}.
The equivalence classes of the above equivalence relation are called
\emph{surreal numbers}. It is possible to define addition and multiplication
on them in such a way that, along with the ordering $<$ already defined,
the surreal numbers satisfy the axioms of an ordered field.

\begin{example}
    (Examples of surreal numbers)
    \begin{itemize}
        \item
        Taking $L=R=\emptyset$ yields the surreal number named $\{\emptyset|\emptyset\}$,
        usually abbreviated $\{|\}$. It can be shown this is the
        surreal $0$ (i.e., the unique surreal additive identity).
        \item
        Let $L=\{0\}$, $R=\emptyset$, where $0$ is as above.
        This yields the surreal named $\{\{0\}|\}$.
        It can be shown this is the surreal $1$ (i.e.,
        the unique multiplicative identity).
        \item
        Let $L=\{1\}$, $R=\emptyset$, where $1$ is as above. This yields the surreal
        number named $\{\{1\}|\}$. It can be shown this is $2$ (i.e., $1+1$).
        \item
        In the same way, one can obtain surreal numbers $3,4,5,\ldots$.
        \item
        Let $L=\{0,1,2,\ldots\}$ (as above), $R=\emptyset$. This yields the surreal
        named $\{\{0,1,2,\ldots\}|\}$. This surreal
        represents the smallest infinite ordinal number, $\omega$, considered as a surreal.
        \item
        With $0$ and $1$ as above, $\{\{0\}|\{1\}\}$ names a surreal strictly between $0$ and
        $1$. It can be shown to be $\frac12$
        (i.e., that its product with $2$ is $1$).
        \item
        With $0$ and $\frac12$ as above, $\{\{0\}|\{\frac12\}\}$ names a surreal strictly
        between $0$ and $\frac12$. It can be shown to be $\frac14$.
        \item
        Similarly, one can construct surreals $\frac18,\frac1{16},\ldots$.
        \item
        With $0$ and $\frac12,\frac14,\frac18,\ldots$ as above,
        $\{\{0\}|\{\frac12,\frac14,\frac18,\ldots\}\}$ names a surreal larger than $0$
        but smaller than every $\frac12,\frac14,\frac18,\ldots$. This shows that the
        surreals include infinitesimals. This particular infinitesimal can be shown
        to be $1/\omega$, in the sense that when multiplied by $\omega$, the result is $1$.
    \end{itemize}
\end{example}

The reader might object that if we let $L$ be the set of all surreals, then
$\{L|\}$ seems to name a surreal larger than all surreals,
which would be absurd. This paradox is avoided because in fact the class
of all surreals is \emph{not} a set, but a proper class\footnote{It can be shown
that the ordinal numbers can be embedded in the surreals, and so the non-sethood
of the class of surreals is strictly weaker than the non-sethood of the class of
ordinals. The latter non-sethood is referred to as the Burali-Forti paradox.}.

It can be shown that for any free ultrafilter $\mathcal U$, the hyperreals constructed using
$\mathcal U$ can be embedded into the surreals. This allows us to transform our hyperreal
solution to Problem \ref{bigoproblem} into a surreal solution.
Unfortunately, there are many different ways to embed the hyperreals into the surreals,
none of them canonical, so while our hyperreal solution already depends arbitrarily on
a free ultrafiler, our surreal solution further depends on an arbitrary embedding.

\begin{definition}
    (The surreal solution to Problem \ref{bigoproblem})
    Suppose $\mathcal U$ is a free ultrafilter and ${}^*\mathbb R$ is the corresponding
    hyperreal number system (Definition \ref{hyperrealsdefn}).
    Let $\iota$ be an embedding of ${}^*\mathbb R$ into the surreals.
    For any function $f:\mathbb N\to\mathbb N$, the \emph{surreal growth rate
    of $f$} (given by $\mathcal U$ and $\iota$) is
    $\iota(\hat f)$ where $\hat f$ is the hyperreal growth
    rate of $f$ (Definition \ref{hyperrealgrowthratedefn}).
\end{definition}

The payoff of doing this is that there is a canonical way to pick a particular
bound of a bounded set of surreals, so the surreal solution to Problem \ref{bigoproblem}
provides an ASPI measure, not just an ASPI taxonomy.

\begin{definition}
\label{surrealhibbardintelligencedefn}
    (The surreal ASPI measure)
    Let $\mathcal U$ be a free ultrafilter, let ${}^*\mathbb R$ be the corresponding
    hyperreal number system, and let $\iota$ be an embedding of ${}^*\mathbb R$
    into the surreals. Let $\iota({}^*\mathbb R)$ be the range of $\iota$.
    For every predictor $p$, the \emph{surreal ASPI measure} of $p$
    (given by $\mathcal U$ and $\iota$)
    is defined to be the surreal with name $\{L|\}$,
    where $L$ is the set of all surreal numbers $\ell\in\iota({}^*\mathbb R)$ such that
    the following condition holds: for every evader $e$,
    if the surreal growth rate of $t_e$ is $<\ell$, then $p$ $\mathcal U$-learns $e$.
\end{definition}

Note that in Definition \ref{surrealhibbardintelligencedefn}, we require
$\ell\in\iota({}^*\mathbb R)$ because otherwise $L$ would not be a set.

\section{Pros and cons of different ASPI measures and taxonomies}
\label{prosandconssection}

Here are pros and cons of the ASPI measures and taxonomies which arise
from different solutions
to the problem (Problem \ref{bigoproblem}) of measuring the growth rate of functions.

\begin{itemize}
    \item
    The original Hibbard measure (Definition \ref{classichibbardmeasuredefn}),
    which arises by measuring growth rate by comparing
    a function with Liu's enumeration \citep{liu1960enumeration} of the primitive
    recursive functions:
    \begin{itemize}
        \item
        Pro: Relatively concrete.
        \item
        Pro: Measures intelligence using a familiar number system (the natural numbers).
        \item
        Con: The numbers which the measure outputs are not very meaningful, in
        that predictor $p$ having a measure of
        $+1$ higher than predictor $q$ tells us little
        about how \emph{much} more computationally complex the evaders which $p$
        learns are, versus the evaders which $q$ learns.
        \item
        Con: Only distinguishes sufficiently non-intelligent predictors; all predictors
        sufficiently intelligent receive measure $\infty$.
    \end{itemize}
    \item
    Big-O/Big-$\Theta$ (Definition \ref{bigointelligencedefn}),
    in which, rather than directly measuring the intelligence of a predictor, instead, we
    would talk of a predictor's intelligence being $O(f(n))$ or $\Theta(f(n))$
    for various functions $f:\mathbb N\to\mathbb N$:
    \begin{itemize}
        \item
        Pro: Nearly perfect granularity (slightly coarser than perfect granularity because
        of the constants $C,C'$ in Definition \ref{bigodefn}).
        \item
        Pro: Computer scientists already use Big-O/Big-$\Theta$ routinely
        and are comfortable with them.
        \item
        Con: A non-numerical taxonomy.
    \end{itemize}
    \item
    Intelligence based on a majorization hierarchy such as the
    standard slow- or fast-growing hierarchy up to $\epsilon_0$
    (Definitions \ref{tradmajorizationhierarchyhibbardmeasuredefn}
    and \ref{fastmajorizationhierarchyhibbardmeasuredefn}):
    \begin{itemize}
        \item
        Pro: A numerical measure, albeit less granular than the
        Big-O/Big-$\Theta$ taxonomies.
        \item
        Pro: Relatively concrete.
        \item
        Pro: The numbers which the measure outputs are meaningful, in the sense that
        the degree to which a predictor $p$ is more intelligent than a
        predictor $q$ is reflected
        in the degree to which $p$'s intelligence-measure is larger than $q$'s.
        \item
        Con: The numbers which the measure outputs are ordinal numbers, which may be
        unfamiliar to some users.
        \item
        Con: Only distinguishes sufficiently non-intelligent predictors; for any particular
        majorization hierarchy, all predictors
        sufficiently intelligent receive measure $\infty$.
    \end{itemize}
    \item
    Hyperreal intelligence (Definition \ref{hyperrealhibbardintelligencedefn}):
    \begin{itemize}
        \item
        Pro: A taxonomy like Big-O/Big-$\Theta$, but with the added benefit
        that the taxons are numerical.
        \item
        Pro: Perfect granularity.
        \item
        Con: Depends on a free ultrafilter (free ultrafilters exist but cannot be
        concretely exhibited).
    \end{itemize}
    \item
    Surreal intelligence (Definition \ref{surrealhibbardintelligencedefn}):
    \begin{itemize}
        \item
        Pro: An actual numerical measure (not just a taxonomy), with
        perfect granularity.
        \item
        Con: The numbers which the measure outputs are surreal numbers,
        which are relatively new and thus unfamiliar, and are difficult
        to work with in practice.
        \item
        Con: Depends on both a free ultrafilter and also an embedding of
        the resulting hyperreals into the surreals.
    \end{itemize}
\end{itemize}

\section{Conclusion}
\label{conclusionsection}

To summarize:
\begin{itemize}
    \item
    \citet{hibbard} proposed an intelligence measure for predictors
    in games of adversarial sequence prediction.
    \item
    We argued that Hibbard's idea actually splits into two orthogonal sub-ideas.
    First: that intelligence can be measured via the growth-rates of the run-times
    of evaders that a predictor can learn to predict. Second: that such growth-rates can
    be measured in one specific way (involving an enumeration of the primitive recursive
    functions). We argued that there are many other ways to measure growth-rates,
    and that each method of measuring growth-rates yields a corresponding
    adversarial sequence prediction intelligence (ASPI) measure or taxonomy.
    \item
    We considered several specific ways of measuring growth-rates of functions, and exhibited
    corresponding ASPI measures and taxonomies. The growth-rate-measuring methods
    which we considered were: Big-O/Big-$\Theta$ notation; majorization hierarchies;
    hyperreal numbers; and surreal numbers.
    \item
    We also discussed how the intelligence of adversarial sequence predictors
    can be considered as an approximation of the intelligence of idealized AGIs.
\end{itemize}

\section*{Acknowledgments}

We acknowledge Bryan Dawson for feedback on Section \ref{hyperrealsubsection}.
We acknowledge Philip Ehrlich for correcting a mistake.
We acknowledge Mikhail Katz and Roman Yampolskiy for providing literature references.
We acknowledge the editor and the reviewers for much generous feedback and suggestions.

\bibliography{hibbard}

\begin{thebibliography}{}

\bibitem[\protect\citeauthoryear{Alexander}{2019a}]{alexander2019intelligence}
Alexander, S.~A.
\newblock 2019a.
\newblock Intelligence via ultrafilters: structural properties of some
  intelligence comparators of deterministic {L}egg-{H}utter agents.
\newblock {\em Journal of Artificial General Intelligence} 10(1):24--45.

\bibitem[\protect\citeauthoryear{Alexander}{2019b}]{ioi1}
Alexander, S.~A.
\newblock 2019b.
\newblock Measuring the intelligence of an idealized mechanical knowing agent.
\newblock In {\em CIFMA}.

\bibitem[\protect\citeauthoryear{Alexander}{2020a}]{ioi2}
Alexander, S.~A.
\newblock 2020a.
\newblock {AGI} and the {Knight-Darwin} Law: why idealized {AGI} reproduction
  requires collaboration.
\newblock In {\em ICAGI}.

\bibitem[\protect\citeauthoryear{Alexander}{2020b}]{alexander2020arch}
Alexander, S.~A.
\newblock 2020b.
\newblock The {A}rchimedean trap: Why traditional reinforcement learning will
  probably not yield AGI.
\newblock {\em Journal of Artificial General Intelligence} 11(1):70--85.

\bibitem[\protect\citeauthoryear{Bostrom}{2003}]{bostrom2003ethical}
Bostrom, N.
\newblock 2003.
\newblock Ethical issues in advanced artificial intelligence.
\newblock In Schneider, S., ed., {\em Science fiction and philosophy: from time
  travel to superintelligence}. John Wiley and Sons.
\newblock  277--284.

\bibitem[\protect\citeauthoryear{Chaitin}{2011}]{chaitin}
Chaitin, G.
\newblock 2011.
\newblock Metaphysics, Metamathematics and Metabiology.
\newblock In Hector, Z., ed., {\em Randomness through computation}. World
  Scientific.

\bibitem[\protect\citeauthoryear{Conway}{2000}]{conway}
Conway, J.~H.
\newblock 2000.
\newblock {\em On Numbers and Games}.
\newblock CRC Press, 2nd edition.

\bibitem[\protect\citeauthoryear{Ehrlich}{2012}]{ehrlich2012absolute}
Ehrlich, P.
\newblock 2012.
\newblock The absolute arithmetic continuum and the unification of all numbers
  great and small.
\newblock {\em Bulletin of Symbolic Logic} 18:1--45.

\bibitem[\protect\citeauthoryear{Girard}{1981}]{girard1981pi12}
Girard, J.-Y.
\newblock 1981.
\newblock {$\Pi^1_2$}-logic, Part 1: Dilators.
\newblock {\em Annals of Mathematical Logic} 21(2-3):75--219.

\bibitem[\protect\citeauthoryear{Goldblatt}{2012}]{goldblatt2012lectures}
Goldblatt, R.
\newblock 2012.
\newblock {\em Lectures on the hyperreals: an introduction to nonstandard
  analysis}.
\newblock Springer.

\bibitem[\protect\citeauthoryear{Good}{1969}]{good1969godel}
Good, I.~J.
\newblock 1969.
\newblock G{\"o}del's theorem is a red herring.
\newblock {\em The British Journal for the Philosophy of Science}
  19(4):357--358.

\bibitem[\protect\citeauthoryear{Hardy}{1904}]{hardy1904theorem}
Hardy, G.~H.
\newblock 1904.
\newblock A theorem concerning the infinite cardinal numbers.
\newblock {\em Quarterly Journal of Mathematics} 35:87--94.

\bibitem[\protect\citeauthoryear{Hibbard}{2008}]{hibbard2008adversarial}
Hibbard, B.
\newblock 2008.
\newblock Adversarial sequence prediction.
\newblock In {\em ICAGI},  399--403.

\bibitem[\protect\citeauthoryear{Hibbard}{2011}]{hibbard}
Hibbard, B.
\newblock 2011.
\newblock Measuring agent intelligence via hierarchies of environments.
\newblock In {\em ICAGI},  303--308.

\bibitem[\protect\citeauthoryear{Hrbacek and Katz}{2020}]{katz}
Hrbacek, K., and Katz, M.~G.
\newblock 2020.
\newblock Infinitesimal analysis without the axiom of choice.
\newblock {\em Preprint}.

\bibitem[\protect\citeauthoryear{Hutter}{2004}]{hutter2004universal}
Hutter, M.
\newblock 2004.
\newblock {\em Universal artificial intelligence: Sequential decisions based on
  algorithmic probability}.
\newblock Springer.

\bibitem[\protect\citeauthoryear{Kirman and Sondermann}{1972}]{kirman}
Kirman, A.~P., and Sondermann, D.
\newblock 1972.
\newblock Arrow's theorem, many agents, and invisible dictators.
\newblock {\em Journal of Economic Theory} 5(2):267--277.

\bibitem[\protect\citeauthoryear{Knuth}{1974}]{knuth}
Knuth, D.~E.
\newblock 1974.
\newblock {\em Surreal numbers: a mathematical novelette}.
\newblock Addison-Wesley.

\bibitem[\protect\citeauthoryear{Knuth}{1976}]{knuth1976big}
Knuth, D.~E.
\newblock 1976.
\newblock Big {Omicron} and big {Omega} and big {Theta}.
\newblock {\em ACM Sigact News} 8(2):18--24.

\bibitem[\protect\citeauthoryear{Legg}{2006}]{legg2006there}
Legg, S.
\newblock 2006.
\newblock Is there an elegant universal theory of prediction?
\newblock In {\em International Conference on Algorithmic Learning Theory},
  274--287.
\newblock Springer.

\bibitem[\protect\citeauthoryear{Liu}{1960}]{liu1960enumeration}
Liu, S.-C.
\newblock 1960.
\newblock An enumeration of the primitive recursive functions without
  repetition.
\newblock {\em Tohoku Mathematical Journal} 12(3):400--402.

\bibitem[\protect\citeauthoryear{Robinson}{1974}]{robinson}
Robinson, A.
\newblock 1974.
\newblock {\em Non-standard analysis}.
\newblock Princeton University Press.

\bibitem[\protect\citeauthoryear{Wainer and
  Buchholz}{1987}]{wainer1987provably}
Wainer, S., and Buchholz, W.
\newblock 1987.
\newblock Provably computable functions and the fast growing hierarchy.
\newblock In Simpson, S.~G., ed., {\em Logic and Combinatorics}. AMS.

\bibitem[\protect\citeauthoryear{Wainer}{1989}]{wainer1989slow}
Wainer, S.
\newblock 1989.
\newblock Slow growing versus fast growing.
\newblock {\em The Journal of Symbolic Logic} 54(2):608--614.

\bibitem[\protect\citeauthoryear{Wang}{2019}]{wang2019defining}
Wang, P.
\newblock 2019.
\newblock On Defining Artificial Intelligence.
\newblock {\em Journal of Artificial General Intelligence} 10(2):1--37.

\bibitem[\protect\citeauthoryear{Weiermann}{1997}]{weiermann1997sometimes}
Weiermann, A.
\newblock 1997.
\newblock Sometimes slow growing is fast growing.
\newblock {\em Annals of Pure and Applied Logic} 90(1-3):91--99.

\bibitem[\protect\citeauthoryear{Weiermann}{2002}]{weiermann2002slow}
Weiermann, A.
\newblock 2002.
\newblock Slow versus fast growing.
\newblock {\em Synthese} 133:13--29.

\bibitem[\protect\citeauthoryear{Yampolskiy}{2012}]{yampolskiy2012ai}
Yampolskiy, R.~V.
\newblock 2012.
\newblock {AI}-complete, {AI}-hard, or {AI}-easy--classification of problems in
  {AI}.
\newblock In {\em The 23rd Midwest Artificial Intelligence and Cognitive
  Science Conference}.

\bibitem[\protect\citeauthoryear{Yampolskiy}{2013}]{yampolskiy2013turing}
Yampolskiy, R.~V.
\newblock 2013.
\newblock Turing test as a defining feature of {AI}-completeness.
\newblock In {\em Artificial intelligence, evolutionary computing and
  metaheuristics}. Springer.
\newblock  3--17.

\bibitem[\protect\citeauthoryear{Yampolskiy}{2020}]{yampolskiycontrol}
Yampolskiy, R.~V.
\newblock 2020.
\newblock On Controllability of Artificial Intelligence.
\newblock {\em Technical report}.

\end{thebibliography}
\end{document}